\documentclass[letterpaper]{article} 
\usepackage{aaai2026}  
\usepackage{times}  
\usepackage{helvet}  
\usepackage{courier}  
\usepackage[hyphens]{url}  
\usepackage{graphicx} 
\usepackage{amsmath}
\usepackage{amssymb}
\usepackage{mathtools}
\usepackage{amsthm}
\usepackage{booktabs} 
\usepackage{subfig}
\usepackage{multirow}
\usepackage{pifont}
\usepackage{natbib}  
\usepackage{caption} 
\usepackage{algorithm}
\usepackage{algorithmic}

\usepackage{newfloat}
\usepackage{listings}
\DeclareCaptionStyle{ruled}{labelfont=normalfont,labelsep=colon,strut=off} 
\floatstyle{ruled}
\newfloat{listing}{tb}{lst}{}
\floatname{listing}{Listing}
\nocopyright  

\title{Convolutionally Low-Rank Models with Modified Quantile\\Regression for Interval Time Series Forecasting}
\author{
    Miaoxuan Zhu\equalcontrib,
    Yi Yu\equalcontrib,
    Yuyang Li,
    Wei Li,
    Guangcan Liu\thanks{Corresponding author.}\\
}  
    
\affiliations{

    Southeast University
}

\usepackage{bibentry}

\begin{document}

\maketitle

\begin{abstract}
The quantification of uncertainty in prediction models is crucial for reliable decision-making, yet remains a significant challenge. Interval time series forecasting offers a principled solution to this problem by providing \textit{prediction intervals} (PIs), which indicates the probability that the true value falls within the predicted range. We consider a recently established \textit{point forecasts} (PFs) method termed Learning-Based Convolution Nuclear Norm Minimization (LbCNNM), which directly generates multi-step ahead forecasts by leveraging the convolutional low-rankness property derived from training data. While theoretically complete and empirically effective, LbCNNM lacks inherent uncertainty estimation capabilities—a limitation shared by many advanced forecasting methods. To resolve the issue, we modify the well-known Quantile Regression (QR) and integrate it into LbCNNM, resulting in a novel interval forecasting method termed LbCNNM with Modified Quantile Regression (LbCNNM-MQR). In addition, we devise interval calibration techniques to further improve the accuracy of PIs. Extensive experiments on over 100,000 real-world time series demonstrate the superior performance of LbCNNM-MQR.
\end{abstract}


\section{Introduction}

Time series forecasting~\cite{de200625}, the problem of making forecasts for future based on historical observations, has found tremendous significance in many areas such as economics~\cite{tang2022survey}, energy~\cite{xie2024novel}, traffic~\cite{chen2024similarity} and meteorology~\cite{wu2023interpretable}. In this paper, we focus on \textit{univariate time series forecasting}, which is to predict the next $h$ unseen values of a given sequence of length $l$, where $h$ is called the \textit{forecast horizon}. We concentrate on univariate time series for two fundamental reasons: (1) they represent the most elementary form of temporal data, serving as the building block for more complex variants such as multivariate~\cite{Shibo_Zhao_Liu_Wu_Shen_2025} and spatio-temporal~\cite{Cao_Wang_Jiang_Yu_Dong_2025} series; (2) they encapsulate the core temporal patterns that underpin effective forecasting models. In general, forecasting is an ill-posed inverse problem, and thus one essentially needs to trade-off between \textit{structural risk} and \textit{empirical risk}:
\begin{equation}
    \label{eq:fmwk}
     \min_f \mathcal{R}\left(f\right)+\lambda\sum_{i=1}^{l} \ell\left(\hat{y}_i,y_i\right),
\end{equation}
where $f$ represents the prediction function that uses historical observations as inputs to generate predictions for all values, including both observed and future values. The prediction $\hat{y}_i$, produced by $f$, corresponds to the observation $y_i$. \(\mathcal{R}(f)\) measures the structural risk brought by $f$, and $\sum_{i=1}^{l}\ell\left(\hat{y}_i,y_i\right)$ quantifies the empirical risk of fitting $f$ to the observations. Under the context of \textit{point forecasts} (PFs), the loss function $\ell\left(\cdot,\cdot\right)$ is often chosen as the Mean Squared Error (MSE) $\ell\left(\hat{y}_i,y_i\right)=(\hat{y}_i-y_i)^2$ or Mean Absolute Error (MAE) $\ell\left(\hat{y}_i,y_i\right)=|\hat{y}_i-y_i|$. In most existing work, $f$ is some parametric model defined explicitly such as Recurrent Neural Networks (RNN)~\cite{lin2023segrnn} or Transformer~\cite{ijcai2024p608}. Since the structural risk is controlled only when designing the prediction function $f$, these methods are prone to overfitting when $f$ is over-parameterized and the training data is little.

Beyond the widely studied PFs mentioned above, there is an increasing recognition of the practical value of \textit{prediction intervals} (PIs). Interval forecasting not only predicts potential future results, but also quantifies the uncertainty inherent in these predictions, providing users with more comprehensive information. This approach is more flexible and robust than PFs, particularly in the face of high uncertainty and dynamic environments, offering valuable support for risk management and decision-making optimization~\cite{fatouros2023deepvar,Yalavarthi_Scholz_Born_Schmidt-Thieme_2025}. However, achieving accurate PIs requires addressing the fundamental trade-off between interval width and coverage probability. For example, a higher coverage requirement may result in excessively wide intervals, which reduce their practicality. Conversely, increasing interval precision may reduce coverage. Identifying the optimal balance point between these two factors is highly challenging.

By and large, PIs could be generated on the basis of PFs. Early interval forecasting methods, e.g., those based on Autoregressive Integrated Moving Average (ARIMA)~\cite{box2015time} and Exponential smoothing~\cite{holt2004forecasting}, assume that the point forecast errors follow certain distributions (e.g., normal distribution) and generate confidence intervals accordingly. Such methods have only limited applicability in real-world scenarios, as the premise of distribution assumption is often invalid. Consequently, researchers establish techniques for constructing distribution-free PIs, mainly including Quantile Regression (QR)~\cite{koenker2001quantile,koenker2017quantile} and Conformal Prediction (CP)~\cite{gibbs2021adaptive,Li_Rodríguez_2025}. In particular, it is quite straightforward for QR to transfer PFs into PIs. Namely, one may replace the loss function in Equation~\ref{eq:fmwk} with the following \textit{quantile loss}:
\begin{equation}
    \label{eq:qr}
    \ell_Q(\hat{y}_i,y_i) = \delta(y_i-\hat{y}_i)^++(1-\delta)(\hat{y}_i-y_i)^+,
\end{equation}
where $(\cdot)^+=\text{max}(0,\cdot)$, and $\delta>0$ is a parameter. Intuitively, when $\delta > 0.5$, the higher cost of underestimating prediction drives the model to generate higher results than the ground truth, and similarity for $\delta < 0.5$. When $\delta = 0.5$, the quantile loss falls back to the MAE loss. In order to generate PIs with a confidence level of $1-\alpha$, the parameter $\delta$ should be set to $1-\alpha/2$ and $\alpha/2$ to calculate the upper and lower bounds of PIs, respectively. Several QR-Based methods have been established~\cite{meinshausen2006quantile,chen2020probabilistic,qr1,qr2}, e.g., Quantile Regression Forests (QRF)~\cite{meinshausen2006quantile} that estimate the conditional distribution of the target variable by building multiple decision trees and aggregating their predictions, DeepTCN~\cite{chen2020probabilistic} that generates multiple prediction results by minimizing aggregated quantile loss at various quantile levels, etc.

To obtain accurate PIs, we would adopt a recently proposed PFs method termed Learning-Based Convolution Nuclear Norm Minimization (LbCNNM)~\cite{liu2022time} as the backbone. Unlike other methods, LbCNNM does not manipulate explicitly the prediciton function, but instead recovers the entire series (including the historical and future parts) from the given observations. The balance between structural risk and empirical risk is determined dynamically and depends on the characteristics of the present series, enabling the possibility of LbCNNM to work well even when only few observations are available for training. Nevertheless, LbCNNM is specific to PFs and incapable of generating PIs. To bridge this gap, we first modify the process of QR with the purpose of smoothing the QR procedure, then integrate it into LbCNNM, resulting in a new method termed LbCNNM with modified QR (LbCNNM-MQR) for interval forecasting. Moreover, we devise interval calibration techniques to further improve the accuracy of PIs. We conduct extensive experiments on over 100,000 real-world time series from different datasets, and the results demonstrate the superior performance of LbCNNM-MQR. In summary, the contributions of this paper mainly include:
\begin{itemize}
    \item We propose a novel technique to smoothen the widely-used QR procedure, which dramatically boosts the prediction performance of the proposed interval forecasting method. This may own independent interests outside the scope of this paper.
    \item We devise a interval forecasting method called LbCNNM-MRQ, which integrates the modified QR technique into LbCNNM. Empirical results demonstrate that LbCNNM-MRQ achieves superior performance on different datasets, which consists of more than 100,000 real-world time series from multiple sources and fields.
\end{itemize}

\section{Notations and Preliminaries}

\subsection{Summary of Main Notations}

Matrices and vectors are denoted as capital letters and bold lowercase letters, respectively. Single numbers are denoted by either lowercase or Greek letters. Three types of matrix norms are used: the Frobenius norm $\Vert \cdot \Vert_F$ defined as the square root of the sum of the squares of the entries of a matrix, the $\ell_1$ norm $\Vert \cdot \Vert_1$ given by the sum of the absolute values of the matrix entries, and the nuclear norm $\Vert \cdot \Vert_*$ calculated as the sum of singular values. For a vector $\mathbf{z}$, $\Vert \mathbf{z} \Vert_1$ and $\Vert \mathbf{z} \Vert_2$ are its $\ell_1$ and $\ell_2$ norms, respectively.



\subsection{Convolutionally Low-Rank Models}
\label{sec:lbcnnm}

For the completeness of presentation, we shall briefly introduce LbCNNM~\cite{liu2022time} and its predecessor called Convolution Nuclear Norm Minimization (CNNM)~\cite{liu2022recovery}.

Different from the majority of forecasting methods, CNNM and LbCNNM approach the problem from the perspective of \textit{compressed sensing}~\cite{donoho2006compressed,candes2008introduction}, regarding forecasting as a special case of \textit{vector completion with arbitrary sampling}. Namely, let $\mathbf{y}=\{y_i\}_{t=1}^m\in\mathbb{R}^m$ be a vector that represents some time series of length $m$ and $\Omega$ be a sampling set consisting of positive integers arbitrarily selected from the range between $1$ and $m$, then the goal is to recover $\mathbf{y}$ based on the observations $\{y_i, i\in\Omega\}$. In this way, it is easy to see that $\Omega=\{1,2,...,m-h\}$ corresponds to the standard forecasting problem of predicting the next $h$ unseen values of a given sequence of length $m - h$. In CNNM and LbCNNM, the number $m$ is a hyper-parameter and called \textit{model size}. When given a sequence $\mathbf{\tilde{y}}\in\mathbb{R}^l$, one should set $m\leq{}l$ and form the observed part of $\mathbf{y}$ by taking the last $m-h$ observations from $\mathbf{\tilde{y}}$.

Given the sampling set $\Omega$ and the observations $\{y_i, i\in\Omega\}$, CNNM recovers the target vector $\mathbf{y}$ by
\begin{equation}
\label{eq:CNNM}
    \min _{\mathbf{x} \in \mathbb{R}^m}\left\|\mathcal{A}_k(\mathbf{x})\right\|_*+\sum_{i \in \Omega}\frac{\lambda k}{2}(x_i-y_i)^2,
\end{equation}
where $\mathcal{A}_k(\cdot)$ is a linear map from $\mathbb{R}^m$ to $\mathbb{R}^{m \times k}$ such that $\mathcal{A}_k(\mathbf{x})$ produces the convolution matrix of $\mathbf{x}$, $x_i$ is the $i$th entry of $\mathbf{x}$, and $k$ is the kernel size used in defining $\mathcal{A}_k(\cdot)$. The term convolution here refers specifically to \textit{circular convolution}, i.e., convolution with the \textit{circulant boundary condition}~\cite{fahmy2012new}. Specifically, $\mathcal{A}_k(\mathbf{x})$ is a truncated version consisting of the first $k$ columns of the circulant matrix generated from $\mathbf{x}$.

Remarkably,~\cite{liu2022recovery} proved that CNNM can exactly recover $\mathbf{y}$ from its observed part, provided that the target $\mathbf{y}$ is convolutionally low-rank. However, the premise of convolutional low-rankness is often not true in practice. To overcome this limitation, \cite{liu2022time} proved that for any vector \( \mathbf{z} \in \mathbb{R}^m \) there always exists an column-wisely orthogonal matrix \( A \in \mathbb{R}^{q \times m} (q \ge m) \) such that \( A\mathbf{z} \) is convolutionally low-rank, and further proposed Learning-Based CNNM (LbCNNM) that performs recovery by
\begin{equation}\label{eq:lbcnnm}
    \min_{\mathbf{x} \in \mathbb{R}^m}\left\|\mathcal{A}_k(A\mathbf{x})\right\|_*+\sum_{i \in \Omega}\frac{\lambda k}{2}(x_i-y_i)^2,
\end{equation}
where $A \in \mathbb{R}^{q \times m}$ that satisfies $A^TA=\mathrm{I}_m$ is a transform matrix learned from $\mathbf{\tilde{y}}$ in advance, $\mathrm{I}_m$ is the $m\times m$ identity matrix, the convolution matrix $\mathcal{A}_k(A\mathbf{x})$ is of size $q\times k$, and the parameter setting of $q=2m$, $k=0.5q$ and $\lambda=1000$ is near-optimal~\cite{liu2022time}. 

Notice that the convex program in Equation~\ref{eq:lbcnnm} defines implicitly a prediction function parameterized by learnable parameters $A \in \mathbb{R}^{q \times m}$. To learn $A$, one needs to first extract a set of training samples, denoted as $\{\mathbf{y}_i\in\mathbb{R}^m\}_{i=1}^n$, from the given sequence $\mathbf{\tilde{y}}\in\mathbb{R}^l$, and then train the model such that $A\mathbf{y}_i$ (for $1\leq{i}\leq{n}$) is convolutionally low-rank. 

\section{Proposed Method}

We now present the proposed interval time series forecasting method, which is built upon three components, including modified QR, interval estimation and interval calibration.

First of all, we shall formalize the interval forecasting problem studied in this paper. Denote by \( \mathbf{\tilde{y}} \in \mathbb{R}^l \) the observed time series of length \( l \), and let $y_i$ be the $i$th value of $\mathbf{\tilde{y}}$. Given a significance level \( \alpha \in (0,1) \) and a forecast horizon \( h \), the goal is to generate \( h \)-step ahead PIs, denoted as $\{\hat{C}^{1-\alpha}_{t}\}_{t=l+1}^{l+h}$, such that the width of PIs is minimized and the coverage requirement specified in the following is satisfied as closely as possible:
\begin{equation}
    \label{eq:coverage}
    \mathbb{P}\left(y_{t}\in\hat{C}^{1-\alpha}_{t}\right)\geq1-\alpha,\enspace\forall{} l+1\leq{}t\leq{}l+h,
\end{equation}
where $y_{t}$ is the ``ground truth'' of the $(t-l)$th unseen future value of the series.

\subsection{Modified Quantile Regression}
\label{sec:MQR}

Methods based on Quantile Regression (QR) play a significant role in interval time series forecasting. By directly predicting values at specified quantiles (e.g., 5\%, 50\%, 95\%), they can generate prediction intervals or provide complete distributional information, offering an intuitive and efficient solution for interval forecasting.

To generate PIs using QR, as aforementioned, one needs to replace the loss function (e.g., MSE and MAE) used for PFs with the quantile loss in Equation~\ref{eq:qr}. The resulted optimization problem can often be solved via Alternating Direction Method of Multipliers (ADMM)~\cite{gabay1976dual,lin2010augmented}, and in this case the subproblems related to quantile loss all boil down to the following standard form:
\begin{equation}
    \label{eq:model with qr}
    \min_{x\in\mathbb{R}}\frac{1}{2}(x-z)^2+\beta\ell_Q(x,y), \enspace \forall y,z\in\mathbb{R},\beta>0.
\end{equation}

It can be proven that the above optimization problem is strongly convex and has a unique minimizer (see Proposition 1 in the technical appendix) given by
\begin{align}\label{eq:median close-form}
&x^*_Q=\text{median}(x^{(1)},x^{(2)},y),\\
&\textrm{with }x^{(1)}=z+\beta\delta\enspace \text{and} \enspace x^{(2)}=z+\beta(\delta-1).\nonumber
\end{align}

Notice that the \textit{median} function is discontinuous and highly sensitive to small variations in its inputs, which may cause instability in the optimization procedure of ADMM. To address this issue, we propose a heuristic approach of replacing the \textit{median} function with the \textit{mean} (see Lemma 3 in the technical appendix for theoretical intuition) , i.e.,
\begin{equation}\label{eq:model with mqr}
    x^*_{MQ}=\text{mean}(x^{(1)},x^{(2)},y).
\end{equation}

Interestingly, this heuristic approach can improve distinctly the accuracy of PIs, as will be shown in our experiments.

\subsection{Interval Estimation}
\label{sec:IE}

Following the sprits of QR, we may replace the MSE loss in Equation~\ref{eq:lbcnnm} by the quantile loss:
\begin{equation}
    \label{eq:LbCNNM-QR}
    \min_{\mathbf{x}\in\mathbb{R}^m} \left\| \mathcal{A}_k\left(A\mathbf{x}\right) \right\|_* +
    \lambda\sum_{i\in\Omega}\ell_Q(x_i,y_i),
\end{equation}
where \(\lambda > 0\) is a hyper-parameter. By setting the parameter \(\delta\) in \(\ell_Q(x_i, y_i)\) to $1-\alpha/2$ and $\alpha/2$, we can obtain the upper and lower bounds of PIs, respectively.
However, different parameter $\lambda$ may be needed to estimate the upper and lower bounds, which is somewhat cumbersome. For convenience, we only use QR to estimate the upper bound (denoted as $\hat{\mathbf{y}}^U$), and the lower bound is obtained via $\hat{\mathbf{y}}^L = 2\hat{\mathbf{y}}^M - \hat{\mathbf{y}}^U$, where $\hat{\mathbf{y}}^M$ is the PFs produced by LbCNNM. The learning of $A$ and the setting of the parameters $q$ and $k$ are also conducted in the same way as LbCNNM.

The convex optimization problem in Equation~\ref{eq:LbCNNM-QR} can be solved by ADMM. We first convert it to the following equivalent problem:
\begin{equation}
\label{eq:equivalent}
    \min_{\mathbf{x}, Z} \|Z\|_{*} + \lambda\sum_{i\in\Omega}\ell_Q(x_i,y_i) \enspace \text{s.t.} \enspace  \mathcal{A}_k(A \mathbf{x}) = Z.
\end{equation}

Then the ADMM algorithm minimizes the augmented Lagrangian function,
\begin{multline}
\label{eq:h(xz)}
    H\left(\mathbf{x},Z\right)= \left\| Z \right\|_* + \lambda\sum_{i\in\Omega}\ell_Q(x_i,y_i)\\
    +   \left\langle\mathcal{A}_k\left(A\mathbf{x}\right)-Z,W\right\rangle +    \frac{\mu}{2}\left\| \mathcal{A}_k(A\mathbf{x})-Z \right\|_F^2,
\end{multline}
with respect to $\mathbf{x}$ and $Z$, by fixing the other variables and then updating the Lagrange multiplier $W$ and the penalty parameter $\mu$. Namely, while fixing the other variables, the variable $Z$ is updated by
\begin{equation}
  \min_{Z} \left\| Z \right\|_* +     \frac{\mu}{2}\left\| Z - \left(\mathcal{A}_k\left(A\mathbf{x}\right)+    \frac{W}{\mu}\right) \right\|_F^2,
\end{equation}
which is solved via Singular Value Thresholding (SVT)~\cite{cai2010singular}. While fixing the others, the variable $\mathbf{x}$ is updated via
\begin{align}
    \label{eq:objfun}
    &\min_{\mathbf{x}}
    \frac{1}{2} \left\| \mathbf{x}-\mathbf{x}^{g0} \right\|_2^2+
    \frac{\lambda}{\mu{}k}\sum_{i\in\Omega}\ell_Q(x_i,y_i),\\
    &\textrm{with } \mathbf{x}^{g0} = -\frac{ A^T\mathcal{A}_k^*\left( W-\mu Z\right)}{\mu{}k},\nonumber
\end{align}
where $\mathcal{A}_k^*$ is the Hermitian adjoint of $\mathcal{A}_k$. Denote by $\mathbf{x}^*_Q$ the minimizer to the above optimization problem, and let
\begin{align*}
    \mathbf{x}^{g1}=\mathbf{x}^{g0}&+\frac{\lambda}{\mu k}\delta, \enspace \mathbf{x}^{g2}=\mathbf{x}^{g0}+\frac{\lambda}{\mu k}(\delta-1),
\end{align*}

Then the $i$th entry of $\mathbf{x}^*_Q$ is denoted by $x_i^Q$ and given by
\begin{equation}\label{eq:solution of lbcnnm-qr}
    x^Q_i =\left\{\begin{array}{cc}
x^{g0}_i,&i\notin\Omega,\\
\textrm{median}(x^{g1}_i,x^{g2}_i,y_i), &i\in\Omega,\\
\end{array}\right.
\end{equation}
where $x^{g0}_i$, $x^{g1}_i$ and $x^{g2}_i$ are the $i$th entries of $\mathbf{x}^{g0}$, $\mathbf{x}^{g1}$ and $\mathbf{x}^{g2}$, respectively. The Lagrange multipliers $W$ and the penalty $\mu$ are updated in the same way as the standard ADMM. The above procedure is referred to as LbCNNM-QR (see Lemma 2 in the technical appendix for a detailed precedure).

As discussed in Equation~\ref{eq:model with mqr}, it could be beneficial to replace the median function in Equation~\ref{eq:solution of lbcnnm-qr} by the mean, resulting in a new procedure called LbCNNM-MQR:
\begin{equation}\label{eq:solution of lbcnnm-mqr}
    x^{MQ}_i =\left\{\begin{array}{cc}
x^{g0}_i,&i\notin\Omega,\\
\textrm{mean}(x^{g1}_i,x^{g2}_i,y_i), &i\in\Omega,\\
\end{array}\right.
\end{equation}
in which all the other variables, including $x^{g0}_i$, $x^{g1}_i$ and $x^{g2}_i$, are calculated in the same way as Equation~\ref{eq:solution of lbcnnm-qr}.

\begin{algorithm}[t]
    \caption{LbCNNM-MQR Based Interval Forecasting}
    \label{alg:alg1}
    \textbf{Input:} Observed time series $\tilde{\mathbf{y}}\in\mathbb{R}^l$, forecast horizon $h$, significance level $\alpha$.\\
    \textbf{Output:} PIs $\{\hat{C}^{1-\alpha}_{t}\}_{t=l+1}^{l+h}$.
    \begin{algorithmic}[1]
        \STATE Compute the preliminary PIs $\{[\hat{y}_t^L, \hat{y}_t^U]\}_{t=l+1}^{l+h}$ by applying LbCNNM-MQR to $\tilde{\mathbf{y}}\in\mathbb{R}^l$.
        \STATE Split $\tilde{\mathbf{y}}=\{y_i\}_{i=1}^l$ into a training set ${\rm Tr}=\{y_i\}_{i=1}^{l-h}$ and a calibration set ${\rm Cal}=\{y_i\}_{i=l-h+1}^{l}$.
        \STATE Compute PIs of ${\rm Cal}$ by applying LbCNNM-MQR to ${\rm Tr}$, and compute a calibration bias $\Delta$ accordingly.
        \STATE Calibrate the preliminary PIs to $\hat{C}^{1-\alpha}_{t} = [\hat{y}_t^L-\Delta, \hat{y}_t^U+\Delta], l+1\leq{}t\leq{}l+h$.
    \end{algorithmic}
\end{algorithm}

\subsection{Interval Calibration}
\label{sec:calibration}

The LbCNNM-MQR procedure generates a series of preliminary PIs in a form as in the following:
\begin{equation}
[\hat{y}_t^L,\hat{y}_t^U], l+1\leq{}t\leq{}l+h,
\end{equation}
where $\hat{y}_t^L$ and $\hat{y}_t^U$ are the lower and upper predictions of the $(t-l)$th unseen future values of the series. However, there is no theoretical guarantee that the above PIs can cover the ground truth with probability at least $1-\alpha$. To fix the issue, we adopt the idea of Conformal Prediction, which has been shown to offer strong theoretical coverage guarantees, and calibrate the preliminary PIs as follows:
\begin{equation}
\hat{C}^{1-\alpha}_{t}=[\hat{y}_t^L - \Delta,\hat{y}_t^U +\Delta], l+1\leq{}t\leq{}l+h,
\end{equation}
where $\Delta>0$ is a calibration bias. In general, $\Delta$ should be estimated from some calibration set such that, on the calibration set, the resulted PIs can cover the ground truth with probability at least $1-\alpha$.

Specifically, the observed series \( \mathbf{\tilde{y}} \in \mathbb{R}^l \) is divided into a training set ${\rm Tr}=\{y_i\}_{i=1}^{l-h}$ and a calibration set ${\rm Cal}=\{y_i\}_{i=l-h+1}^{l}$, where \( {\rm Cal} \) plays the role of the ground truth for evaluating predictions made based on \( {\rm Tr} \). Then the calibration bias $\Delta$ is calculated as follows. First, obtain PIs of the calibration set, denoted as $\{[\hat{y}_t^{L}, \hat{y}_t^{U}]\}_{t=l-h+1}^{l}$, by applying LbCNNM-MQR to the training set ${\rm Tr}$. Second, define and compute the residual conformity score set \( \mathbf{s}_{\rm Cal} \), with \( \mathbf{s}_{\rm Cal} \) being consist of residuals $s_j^L=|y_j - \hat{y}_j^{L}|$, $s_j^M=|y_j - \hat{y}_j^{M}|$ and $|y_j - \hat{y}_j^{U}|$, $\forall{}l-h+1\leq{}j\leq{}l$. Finally, the calibration bias $\Delta$ is set to the $(1-\alpha)$th quantile of \( \mathbf{s}_{\rm Cal} \). Algorithm~\ref{alg:alg1} summarizes the whole procedure of our LbCNNM-MQR based interval time series forecasting method.

\section{Experiments}
\label{sec:exp}

\subsection{Datasets}

This paper mainly conducts experiments on the M4 dataset~\cite{makridakis2020m4}, which contains 100,000 real-world time series. The M4 dataset covers various domains, including industry, finance, population, transportation, nature, etc., and is categorized into six groups based on sampling frequency: hourly, daily, weekly, monthly, quarterly and yearly. In addition, any information that could possibly lead to the identification of the original series was removed to ensure the objectivity of the results.

We also include the commonly used multivariate time series forecasting datasets Electricity~\cite{electricityloaddiagrams20112014_321} and Traffic~\cite{wu2021autoformer} as supplementary benchmarks. For the univariate time series forecasting task, the Electricity dataset contains 321 time series and the Traffic dataset contains 862 time series, both sampled at an hourly frequency.

The information of the datasets is shown in Table~\ref{tab:dataset-info}. We provide the forecastability~\cite{wang2024timemixer} of all datasets, which is calculated by one minus the entropy of Fourier decomposition of time series~\cite{goerg2013forecastable}. A higher value indicates better predictability. This indicates that the M4 dataset presents a more challenging test.

\begin{table}[t]
\centering
\begin{tabular}{@{}lcccc@{}}
\toprule
Dataset &
  \begin{tabular}[c]{@{}c@{}}Number\\ of series\end{tabular} &
  \begin{tabular}[c]{@{}c@{}}Forecast\\ horizon\end{tabular} &
  Forecastability \\ \midrule
Electricity  & 321    & 96 & 0.77 \\
Traffic      & 862    & 96 & 0.68 \\
M4-Hourly    & 414    & 48 & 0.46 \\
M4-Daily     & 4227   & 14 & 0.44 \\
M4-Weekly    & 359    & 13 & 0.43 \\
M4-Monthly   & 48,000 & 18 & 0.44 \\
M4-Quarterly & 24,000 & 8  & 0.47 \\
M4-Yearly    & 23,000 & 6  & 0.43 \\
\bottomrule
\end{tabular}%
\caption{Dataset Information.}
\label{tab:dataset-info}
\end{table}

\begin{table*}[t]
\centering
\begin{tabular}{@{}c|cc|cc|cc|cc|cc@{}}
\toprule
Dataset &
  \multicolumn{2}{c|}{Yearly} &
  \multicolumn{2}{c|}{Quarterly} &
  \multicolumn{2}{c|}{Monthly} &
  \multicolumn{2}{c|}{Others} &
  \multicolumn{2}{c}{Overall} \\ \midrule
Metric    & MSIS         & ACD             & MSIS            & ACD          & MSIS        & ACD          & MSIS         & ACD             & MSIS         & ACD          \\ \midrule
AutoARIMA & 53.044       & 22.57\%         & 11.936          & 8.15\%       & 9.895       & 4.09\%       & 37.406       & 2.82\%          & 21.685       & 9.25\%       \\
AutoETS   & \underline {35.445} & 10.88\%         & \textbf{10.934} & 3.85\%       & \underline {9.379} & 1.87\%       & 59.954       & \underline {0.57\%}   & \underline {18.276} & 4.35\%       \\
AutoTheta & 52.618       & 25.63\%         & 14.444          & 15.60\%      & 11.304      & 7.66\%       & 36.768 & 5.25\%          & 22.833       & 13.58\%      \\
DeepAR    & 80.941       & 8.12\%          & 32.262          & 25.91\%      & 19.903      & 24.50\%      & 49.665       & 13.23\%         & 38.396       & 20.51\%      \\
DeepTCN   & 68.731       & \textbf{0.46\%} & 19.905          & \underline {0.57\%} & 23.009      & 1.85\% & 48.012       & \textbf{0.52\%} & 34.030       & \underline {0.83\%} \\ \midrule
LbCNNM-CP & 45.533       & 19.00\%         & 14.635          & 15.83\%      & 10.713      & 9.65\%       & 42.217       & 19.54\%         & 21.238       & 13.78\%      \\
LbCNNM-QR & 39.285
       & 11.77\%
        & 12.119
          & 5.90\%
     & 10.078
      & \underline {1.27\%}
      & \underline{34.479}
       & 2.44\%
         & 18.505
       & 4.85\%
     \\
LbCNNM-MQR &
  \textbf{33.046} &
  \underline {1.70\%} &
  \underline {11.297} &
  \textbf{0.44\%} &
  \textbf{9.373} &
  \textbf{0.15\%} &
  \textbf{31.162} &
  0.70\% &
  \textbf{16.369} &
  \textbf{0.18\%} \\ \bottomrule
\end{tabular}%
\caption{Interval forecasting results on the M4 dataset with $\alpha$ = 0.05. A lower MSIS or ACD indicates a better prediction.The best results are highlighted in \textbf{bold} and the second best are \underline{underlined}.}
\label{tab:main-results}
\end{table*}

\subsection{Evaluation Metrics}

In this paper, a 95\% prediction interval (i.e., \( \alpha = 0.05 \)) is used to evaluate the experimental results of interval forecasting. This confidence level is selected because it is one of the most commonly used levels in the business world, as it is neither too tight (e.g. 99\%) nor too wide (e.g. 90\%) for the majority of economic and financial forecasting applications.

To evaluate the performance of PIs generated by the algorithm, this paper adopts the evaluation metrics used in the M4 competition. The first metric is the Mean Scaled Interval Score (MSIS)~\cite{Gneiting01032007}, defined as follows:
\begin{small}
\begin{align}
&\text {MSIS}=\frac{1}{h} \nonumber\\
&\times \!\frac{\sum\!_{t=n+1}^{n+h}\!\!\left(U_{t}\!-\!L_{t}\right)\!+\!\frac{2}{\alpha}\!\!\left(L_{t}\!-\!Y_{t}\right)\! \textbf{1} Y_{t}\!\!<\!\!L_{t}\!+\!\frac{2}{\alpha}\!\!\left(Y_{t}\!-\!U_{t}\right) \!\textbf{1} Y_{t}\!\!>\!\!U_{t}}{\frac{1}{n-m} \sum_{t=m+1}^{n}\left|Y_{t}-Y_{t-m}\right|},  \nonumber
\end{align}
\end{small}

where $L_t$ and $U_t$ are the lower and upper bounds of PIs, $Y_t$ is the value of the time series, $h$ is the forecast horizon, $n$ is the number of the data points available in-sample, $\alpha$ is the significance level and $\textbf{1}$ is the indicator function (being 1 if $Y_t$ is within the generated interval and 0 otherwise). $m$ (here $m$ differs from the usage in convolutionally low-rank models) is the \textit{time interval} between successive observations considered by the organizers for each data frequency, i.e., 24 for hourly, 12 for monthly, 4 for quarterly, and 1 for yearly, weekly and daily data.

For MSIS, the width of PIs is always included as part of the penalty term. In this respect, the methods of larger intervals are penalized over those of smaller ones, regardless of the coverage rate achieved. A penalty is calculated for each method at the points at which the future values are outside the specified bounds. This captures the coverage rate of each method. Finally, the measure is scaled by dividing by the average absolute seasonal variation of the series, making it scale-independent.

The second metric is the Absolute Coverage Difference (ACD), which is simply the absolute difference between the average coverage of the method and the target set. 
For instance, if the future values across the 100,000 time series are outside the bounds specified by a method an average of \(2\%\) of the time (coverage of \(98\%\)), the ACD will be $\left|0.95 - 0.98\right| = 3\%$.


\subsection{Baselines}

We compare LbCNNM-MQR with seven baseline methods, including AutoARIMA~\cite{hyndman2008automatic} and AutoETS~\cite{hyndman2018forecasting}, which automatically select the best model using the Akaike Information Criterion (AICc)~\cite{akaike1974new}, and AutoTheta~\cite{FIORUCCI20161151}, which automatically selects the best model using MSE. Two deep learning prediction methods, DeepAR~\cite{salinas2020deepar} and DeepTCN~\cite{chen2020probabilistic} are also included in the comparison. For comparison, we implement two additional interval forecasting methods based on LbCNNM: one is LbCNNM-QR that employs standard quantile regression in Equation~\ref{eq:median close-form}, the other is LbCNNM-CP that directly combines LbCNNM with Conformal Prediction (CP).

\subsection{Experimental Results}
\label{sec:results}

\subsubsection{Comparison results}
\label{sec:main-results}

\begin{table}[t]
\centering
\begin{tabular}{@{}c|cc|cc@{}}
\toprule
Dataset    & \multicolumn{2}{c|}{Electricity} & \multicolumn{2}{c}{Traffic} \\ \midrule
Metric     & MSIS             & ACD           & MSIS          & ACD         \\ \midrule
AutoARIMA  & 12.287          & 2.43\%        & 19.694       & 0.87\%      \\
AutoETS    & 9.140         & 0.53\%        & 27.227      & 2.59\%      \\
AutoTheta  & 15.725          & 3.93\%        & 31.525      & 4.77\%      \\
DeepAR     & 9.021           & 1.71\%        & 9.992        & 1.59\%      \\
DeepTCN    & 10.042           & 3.71\%        & 11.418        & 3.23\%      \\ \midrule
LbCNNM-MQR & 8.140           & 0.70\%        & 12.200       & 2.53\%      \\ \bottomrule
\end{tabular}%
\caption{Interval forecasting results on the Electricity and Traffic dataset. }
\label{tab:other_dataset}
\end{table}

The comparison of the experimental results between LbCNNM-MQR and the competing methods on the M4 dataset is presented in Table~\ref{tab:main-results}. ``Others'' refers to the aggregation of the hourly, daily, and weekly high-frequency datasets, which collectively comprise a total of 5,000 time series. ``Overall'' represents the entire M4 dataset, consisting of 100,000 time series.

From the experimental results, LbCNNM-MQR demonstrates the best performance on the entire M4 dataset and consistently achieves superior performance across different sampling frequencies. In terms of MSIS, LbCNNM-MQR falls behind AutoETS by 3.3\% only on the quarterly dataset while it achieves a 10.4\% improvement over the second-best method, AutoETS, on the entire M4 dataset. In terms of coverage, which measures the validity of PIs, the average ACD of LbCNNM-MQR on the entire M4 dataset is 0.18\%, exceeds all competing methods. This results indicates that LbCNNM-MQR almost achieves the nominal coverage requirements. 

Compared with DeepTCN, which similarly provides reliable coverage, LbCNNM-MQR outperforms in terms of MSIS across all frequency datasets. This indicates that LbCNNM-MQR does not sacrifice the average width of PIs to maintain coverage, making it more practical and valuable for real-world applications. Compared with LbCNNM-CP, two proposed optimization methods combined with quantile regression show notable progress in both MSIS and ACD. This indicates that the effectiveness of the algorithm is not merely reliant on the basic CP. The overall superiority of LbCNNM-MQR over LbCNNM-QR demonstrates the effectiveness of the proposed modified quantile regression.

Table~\ref{tab:other_dataset} presents the results of LbCNNM-MQR compared with competing methods on the Electricity and Traffic datasets. In our experiments, the first 720 time steps are used as the training data, followed by the next 96 time steps as the test data, with the time interval set to 24. For the two datasets with richer training data and more pronounced periodic patterns, deep learning models exhibit superior performance.

Table~\ref{tab:running time} presents the average running time of different models on the M4 dataset. All experiments were conducted on an Intel(R) Xeon(R) Platinum 8358P CPU platform, with DeepAR and DeepTCN executed on a single NVIDIA A800 80GB GPU.

In general, considering the accuracy of the prediction, LbCNNM-MQR achieves the best balance between performance and efficiency. More experimental details will be shown in the technical appendix.

\begin{table}[t]
\centering
\begin{small}
\setlength{\tabcolsep}{1mm}{
\begin{tabular}{@{}cccccccc@{}}
\toprule
          & Yearly & Quarterly & Monthly & Others  & Overall \\ \midrule
AutoARIMA & 0.169 & 0.239    & 0.669  & 1.098  & 0.472  \\
AutoETS   & 0.012 & 0.022    & 0.027  & 0.429  & 0.042  \\
AutoTheta & 0.012 & 0.063    & 0.119  & 7.523  & 0.451  \\
DeepAR    & 0.620 & 1.172    & 2.870  & 35.493 & 3.576  \\
DeepTCN   & 0.694 & 1.568    & 2.945  & 29.644 & 3.432  \\ \midrule
LbCNNM-MQR  & 0.085 & 0.251    & 1.324  & 4.734 & 0.952  \\ \bottomrule
\end{tabular} }
\end{small}
\caption{Average running time(s) on the M4 dataset.}
\label{tab:running time}
\end{table}

\subsubsection{Experimental Analysis}
\label{sec:ablation}

\begin{figure}[t]
    \centering
    \subfloat[LbCNNM-QR]
    {
        \label{fig:median-lambda}
        \includegraphics[width=0.48\columnwidth]{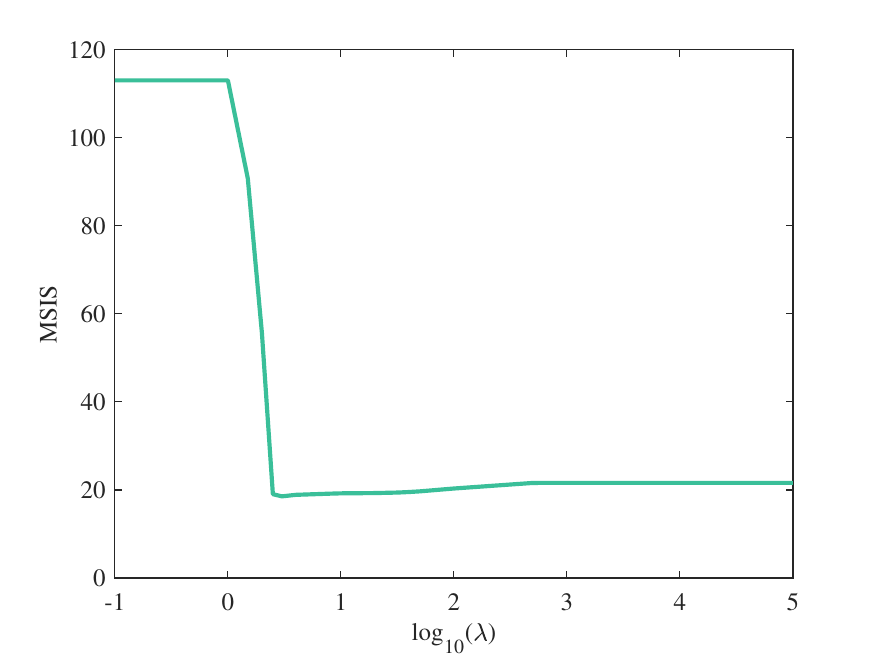}
    }\hspace{-10pt}
    \subfloat[LbCNNM-MQR]
    {
        \label{fig:mean-lambda}
        \includegraphics[width=0.48\columnwidth]{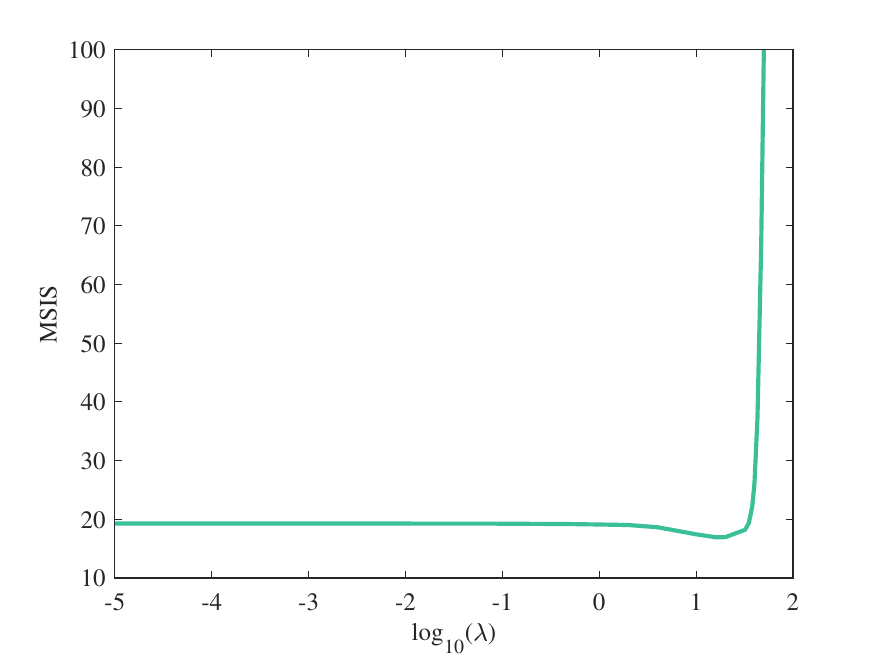}
    }
    \caption{The influence of hyper-parameter $\lambda$ on LbCNNM-QR and LbCNNM-MQR, using proportionally randomly selected M4-small dataset.}
    \label{fig:lambda}
\end{figure}

\begin{figure}[t]
    \centering
    \subfloat[$\lambda=2,\text{LbCNNM-QR}$]
    {
        \label{fig:2a}
        \includegraphics[width=0.48\columnwidth]{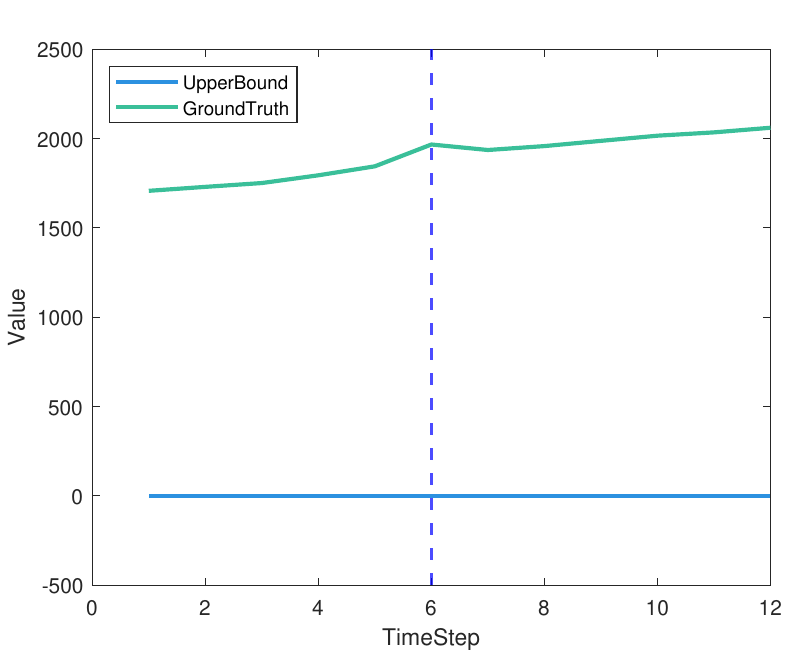}
    }\hspace{-10pt}
    \subfloat[$\lambda=20,\text{LbCNNM-QR}$]
    {
        \label{fig:2b}
        \includegraphics[width=0.48\columnwidth]{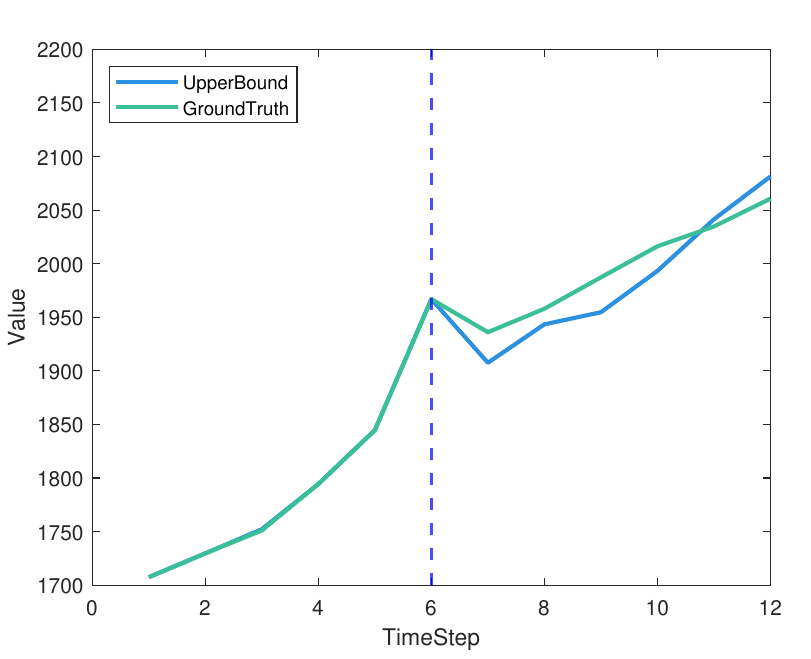}
    }
    \\
    \subfloat[$\lambda=2,\text{LbCNNM-MQR}$]
    {
        \label{fig:2c}
        \includegraphics[width=0.48\columnwidth]{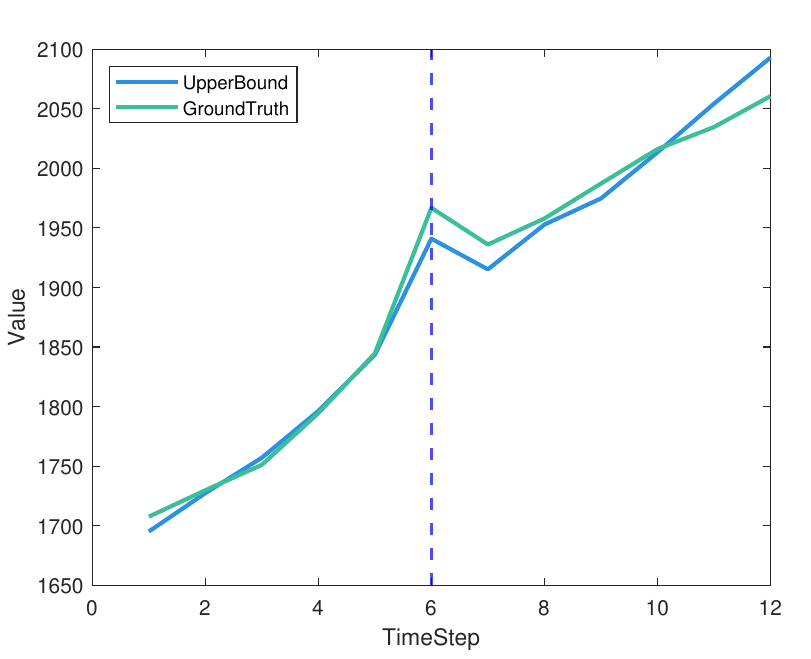}
    }\hspace{-10pt}
    \subfloat[$\lambda=20,\text{LbCNNM-MQR}$]
    {
        \label{fig:2d}
        \includegraphics[width=0.48\columnwidth]{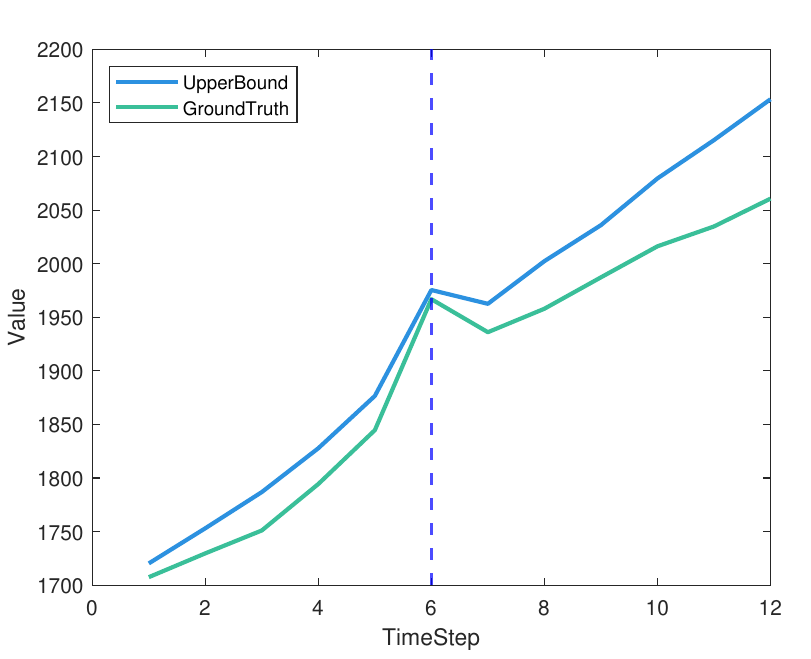}
    }
    \caption{Visualization of PIs upper bounds under of different settings during the interval estimation phase for a yearly series.
}
    \label{fig:fig2}
\end{figure}

In this section, we will analyze the impact of model settings. Except the hyper-parameter \(\lambda\), the other parameters are fixed in the way as suggested by~\cite{liu2022time}. To investigate the influence of $\lambda$, we randomly sample one-tenth of the series from the entire M4 dataset, proportionally to the number of series in each subset (e.g., selecting 4,800 series at random from the monthly dataset), to create a new dataset -- M4-small with 10,000 series. Figure~\ref{fig:lambda} presents the MSIS of LbCNNM-QR and LbCNNM-MQR evaluated on M4-small with different values of \(\lambda\). Figure~\ref{fig:fig2} visualizes the results obtained during the interval estimation phase with different $\lambda$ for a same yearly series. The dashed line separates the observation (left) and prediction (right) parts.

It can be observed from Figure~\ref{fig:lambda} that when \(\lambda\) is above or below a certain threshold, LbCNNM-QR and LbCNNM-MQR can achieve stable and satisfactory results. This suggests that the algorithm shows robustness to hyper-parameter settings and demonstrates strong generalization capability. It can also be seen that the two algorithms exhibit seemingly opposite trends with respect to \(\lambda\). For LbCNNM-QR, this is because when \(\lambda\) is too small, the weight of the quantile loss term in Equation~\ref{eq:LbCNNM-QR} decreases, resulting in low-rank results dominated by the nuclear norm like Figure~\ref{fig:2a}. When \(\lambda\) is too large, the quantile loss term becomes dominant, causing the nuclear norm to have no effect. Since its extremum is at \(\mathbf{y}\), the predictions for the observed part closely coincide with the true values like Figure~\ref{fig:2b}, losing the ability to control PIs according to \(\delta\), which is a problem of standard quantile regression.

\begin{table}[t]
\centering
\begin{scriptsize}
\setlength{\tabcolsep}{0.5mm}{
\begin{tabular}{@{}c|cc|cc|cc@{}}
\toprule
\multirow{2}{*}{Case} &
  \multicolumn{2}{c|}{Interval estimation} &
  \multicolumn{2}{c|}{Interval calibration} &
  \multicolumn{2}{c}{M4 overall} \\ \cmidrule(l){2-7}
 &
  $\mathbf{x}^*_Q$ &
  $\mathbf{x}^*_{MQ}$ &
  $\mathbf{s}_{\rm Cal}=\{s_j^M\}$ &
  $\mathbf{s}_{\rm Cal}=\{s_j^U,s_j^M,s_j^L \}$ &
  MSIS &
  ACD \\ \midrule
\ding{172} & $\checkmark$ & $\times$     & $\checkmark$ & $\times$     & 18.980          & 7.53\%          \\
\ding{173} & $\checkmark$ & $\times$     & $\times$     & $\checkmark$ & 18.505          & 4.85\%          \\
\ding{174} & $\times$     & $\checkmark$ & $\checkmark$ & $\times$     & 16.563          & 3.56\%          \\
\ding{175} & $\times$     & $\checkmark$ & $\times$     & $\checkmark$ & \textbf{16.369} & \textbf{0.18\%} \\ \bottomrule
\end{tabular} }
\end{scriptsize}
\caption{Performance of the algorithm with different settings on the M4 dataset.
}
\label{tab:ablation}
\end{table}

\begin{figure*}[t]
    \centering
    \subfloat[]
    {
        \label{fig:3a}
        \includegraphics[width=0.33\textwidth]{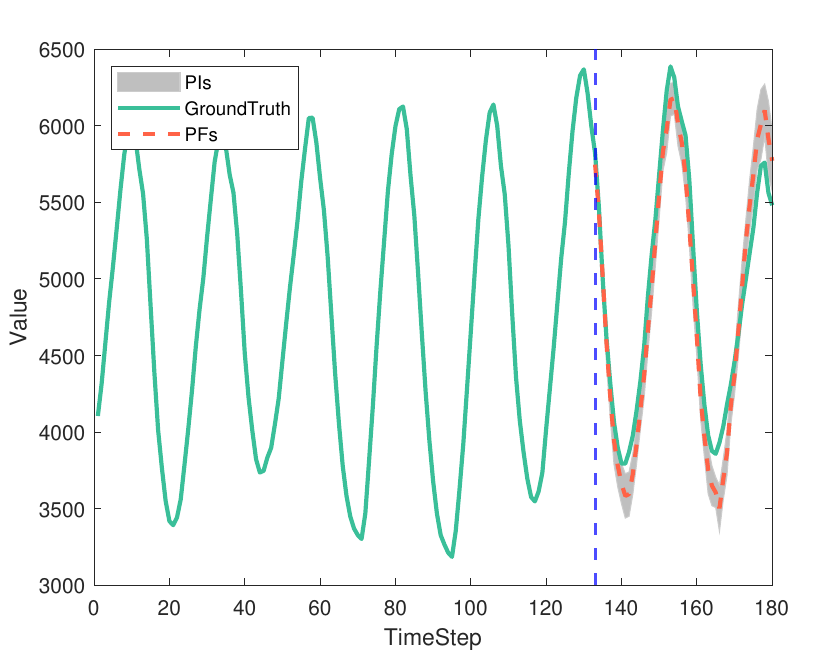}
    }\hspace{-10pt}
    \subfloat[]
    {
        \label{fig:3b}
        \includegraphics[width=0.33\textwidth]{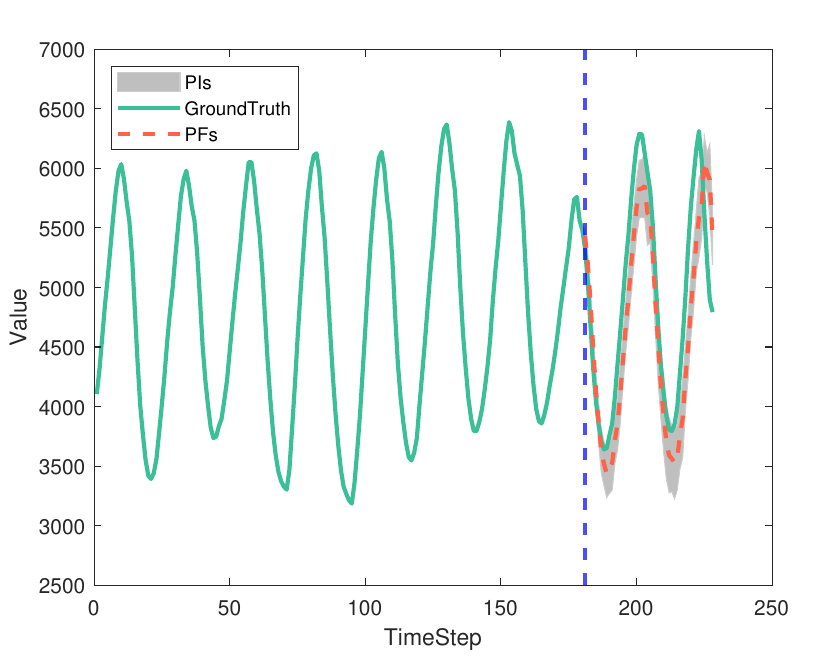}
    }\hspace{-10pt}
    \subfloat[]
    {
        \label{fig:3c}
        \includegraphics[width=0.33\textwidth]{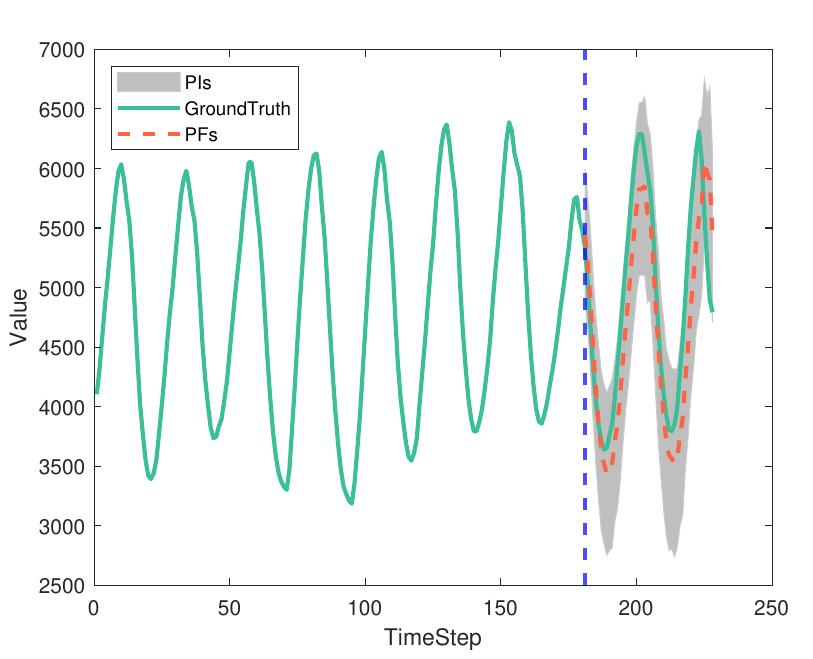}
    }
    \caption{Visualization of PIs at different phases of LbCNNM-MQR for an hourly series. (a) shows the interval estimation results on the calibration set, (b) shows the interval estimation results for the prediction part, and (c) shows the interval calibration results for the prediction part.}
    \label{fig:fig3}
\end{figure*}
For the curve of LbCNNM-MQR with respect to $\lambda$ shown in Figure~\ref{fig:mean-lambda}, we first need to understand the meaning of the solution to Equation~\ref{eq:solution of lbcnnm-mqr}. Substituting ${x}^{g1}_i,{x}^{g2}_i$ into the equation ${x}^{MQ}_i=\text{mean}({x}^{g1}_i,{x}^{g2}_i,{y}_i),i\in\Omega$, we get:
\begin{align*}
\label{eq:mean}
    x^{MQ}_i = \frac{2}{3}x^{g0}_i+\frac{1}{3}y_i+\frac{1}{3}\left(\frac{\lambda(2\delta-1)}{\mu k}\right).
\end{align*}
Note that $x^{g0}_i$ and $y_i$ represent the respective extreme points of the $\ell_2$ loss and the quantile loss in the objective function presented in Equation~\ref{eq:objfun}. For the two convex subfunctions that constitute the objective function, each of them has a unique extreme point, so the extreme point of the objective function obtained by their linear combination must be between their respective extreme points. In other words, the iteration result presented by the modified quantile regression contains the linear interpolation of the two extreme points, the bottom line of the prediction results is ensured, even for a small \(\lambda\) like Figure~\ref{fig:2c}. Additionally, the third term in the above equation makes a critical contribution to PIs. When \( \delta > 0.5 \), i.e., in the case of calculating the upper bound of PIs, the result of this term is positive, indicating that it will shift the result upward. Conversely, for \( \delta < 0.5 \), when calculating the lower bound, it will adjust the result downward, which is consistent with the logical requirements of PIs. Therefore, provided that \(\lambda\) is not set too large, as this would result in excessively wide intervals, LbCNNM-MQR can generate satisfactory PIs like Figure~\ref{fig:2d}.

We explore the influence of the set of conformity score \(\mathbf{s}_{\rm Cal}\) during the interval calibration phase in Table~\ref{tab:ablation}. When comparing the combined case \ding{172}\ding{173} with case \ding{174}\ding{175}, the $\mathbf{x}^*_{MQ}$ solution outperforms in both MSIS and ACD. Compared to the \(\mathbf{s}_{\rm Cal} = \{s_j^M\}\) commonly used in PFs-based methods, \(\mathbf{s}_{\rm Cal}=\{s_j^U,s_j^M,s_j^L\}\), which includes interval boundary residuals, improves the calibration of preliminary PIs from the calibration set, resulting in better coverage in the prediction part. From the perspective of MSIS, which balances both interval width and coverage, these two effects offset each other. Therefore, users can flexibly select the corresponding $\mathbf{s}_{\rm Cal}$ according to the specific goals of the application.

We also visualize the effect of the interval calibration in Figure~\ref{fig:fig3}. Figure~\ref{fig:3a} and Figure~\ref{fig:3b} show the interval estimation results on the calibration set and the prediction part, respectively. Figure~\ref{fig:3c} presents the results after interval calibration. By calculating the residuals on ${\rm Cal}$, PIs that initially could not fully cover the ground truth achieve better forecasting performance.

\section{Conclusion}
In this work, we propose a modified quantile regression with the purpose of smoothing the standard quantile regression procedure. By integrating it with the recently proposed PFs method LbCNNM, we devise a novel and simple algorithm for interval time series forecasting, termed LbCNNM-MQR. Firstly, a closed-form solution is theoretically derived for preliminary interval estimation, directly generating multi-step ahead PIs. Subsequently, we adopt interval calibration techniques to further improve the accuracy of PIs by calculating prediction errors on the calibration set divided from the observed part. Extensive experiments on over 100,000 time series from different datasets demonstrate the superior performance of LbCNNM-MQR, particularly in terms of interval coverage. Meanwhile, LbCNNM-MQR exhibits its practicality in real-world scenarios due to favorable run-time efficiency and applicability to any series without parameter tuning. Comparison experiments with the standard QR-based method LbCNNM-QR confirm the effectiveness of the proposed modified quantile regression, which may own independent interests outside the scope of this paper. 

\bibliography{aaai2026}


\end{document}